\documentclass[times, 10pt,twocolumn]{article}
\usepackage{latex8}
\usepackage{times}
\usepackage{graphicx}
\usepackage{color}
\usepackage{booktabs}
\usepackage{color}

\usepackage{hueifang}

\begin{document}

\title{Joint Estimation of Age and Gender from Unconstrained Face Images \\
using Lightweight Multi-task CNN for Mobile Applications}

\author{Jia-Hong Lee, Yi-Ming Chan, Ting-Yen Chen, and Chu-Song Chen\\
Institute of Information Science, Academia Sinica, Taipei\\
$\{honghenry.lee, yiming, timh20022002, song\}@iis.sinica.edu.tw$
}

\maketitle
\thispagestyle{empty}

\begin{abstract}
Automatic age and gender classification based on unconstrained images has become essential techniques on mobile devices. 
With limited computing power, how to develop a robust system becomes a challenging task.
In this paper, we present an efficient convolutional neural network (CNN) called lightweight multi-task CNN for simultaneous age and gender classification. 
Lightweight multi-task CNN uses depthwise separable convolution to reduce the model size and save the inference time. 
On the public challenging Adience dataset, the accuracy of age and gender classification is better than baseline multi-task CNN methods.

\end{abstract}
\section{Introduction}
Understanding age and gender from the human face plays an essential role in social interaction. 
To make communication proper and efficient, people subconsciously judge others' age or gender.
Thus, age and gender estimation is important in several applications, including re-identification in surveillance videos, intelligent advertising and human-computer interaction. 
Nevertheless, accurately and efficiently estimating age and gender from unconstrained face images is difficult.

Prior to deep neural network era, several approaches estimate age and gender from face images using designed image features and machine learning.
In~\cite{eidinger2014age}, Eidinger \etal combine four-patch local binary patterns (FPLBP)~\cite{wolf2008descriptor} and support vector machine (SVM)~\cite{cortes1995support} to achieve the joint-estimation of age and gender.
Han \etal~\cite{han2015demographic} use biologically inspired features (BIF) and their designed hierarchical estimator for this task.
Since deep CNNs get a great success in object classification, Rothe \etal~\cite{rothe2015dex} develop the DEX method which consists of the face detector in~\cite{mathias2014face} and a deep CNN architecture VGG-16~\cite{simonyan2014very} for age estimation.
Levi\_Hassner~\cite{levi2015age} introduce a five-layer CNN architecture that achieves the most favorable performance on the unconstrained public Adience dataset~\cite{eidinger2014age}.


Although the Levi\_Hassner CNN~\cite{levi2015age} can achieve high accuracy, it needs two independent models for predicting the age and gender, respectively. 
To reduce the memory cost, we develop a method simplifying the weights with a single light weight model through multi-task learning and the technique of depthwise separable convolution.

\section{Related Work} \label{sec:relatedwork}

There are two types of structures commonly used in multi-task learning with deep neural networks~\cite{ruder2017overview}, $hard~parameter~sharing$ and $soft~parameter~sharing$ of hidden layers. 
Soft parameter sharing means that each task has its own deep neural network with same structure, and then a similarity function is utilized to regularize these models~\cite{yang2016trace}. 
Thus the space required at run time is proportional to the number of tasks. 
To  reduce the space complexity, the structure of hard parameter sharing is most commonly used in deep multi-task learning. 
It only employs one shared deep neural network and keeps several task-specific output layers~\cite{zhang2014facial}.
The hard parameter sharing can not only reduce the space complexity, but can also decrease the risk of over-fitting~\cite{baxter1997bayesian}.

\begin{figure}
\centering
\includegraphics[width=8cm,keepaspectratio]{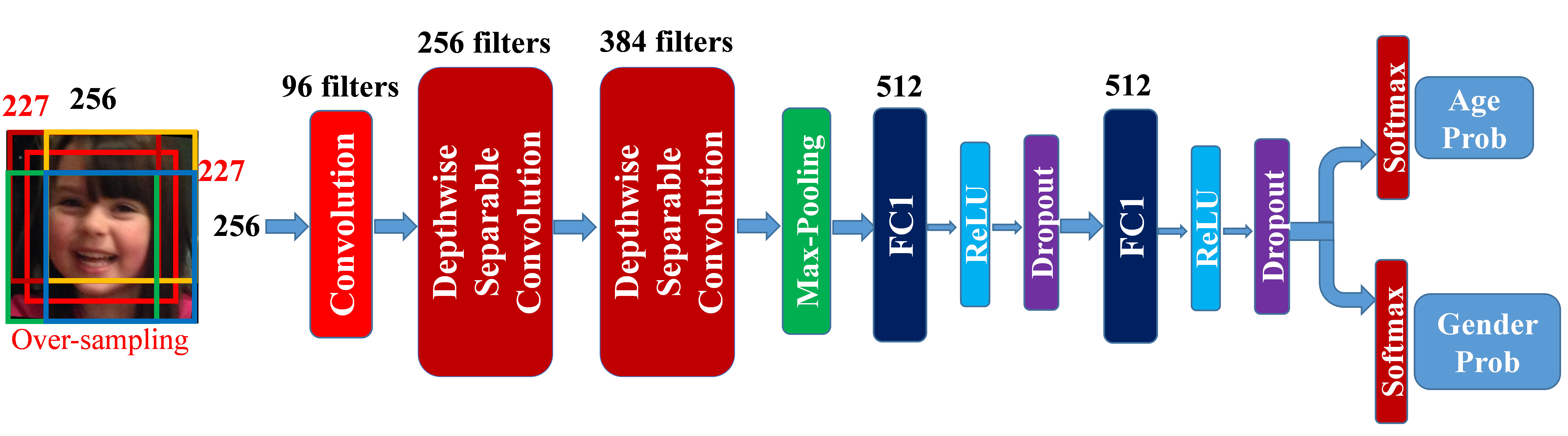}
\caption{The architecture of the LMTCNN.}
\label{fig:lightenedmultitaskcnn}
\end{figure}


The getting popular mobile device motivates researchers to develop deep neural networks for mobile applications~\cite{howard2017mobilenets,li2017deeprebirth,zhang2017shufflenet, kim2015compression}.
MobileNet~\cite{howard2017mobilenets} is one of the most interesting approaches about speedup of a deep neural network. 
The MobileNet is depended on a streamlined architecture that uses $depthwise~separable~convolutions$ to factorize a general convolution into a depthwise convolution and a pointwise convolution. 
By combining the output values of the depthwise convolution with point wise convolution, a lightweight deep neural network thus can be constructed. 
To further parameterize the tradeoff between latency and accuracy, they use two global hyper-parameters, width~multiplier and resolution~multiplier to adjust the computational cost of the neural network.


\section{Lightweight Multi-Task CNN} \label{sec:methodology}

We refine the state-of-the-art approach~\cite{levi2015age} on two aspects: simultaneous inference and model redution, which are critical for mobile applications.
Unlike Levi\_Hassner CNN~\cite{levi2015age} that uses two independent models to recognize age and gender, only one single CNN for feature extraction is used for multiple tasks in our system.
Thus, the memory requirement for deep neural networks is reduced.
We employ a hard parameter sharing paradigm to learn the single CNN for both tasks. 
To further decrease the computation cost, we decompose the general CNN in~\cite{levi2015age} into depthwise and pointwise convolution networks.
The pointwise convolution is a convolution with $1\times1$ kernel's size, and it combines the output values of the depthwise convolution.

\subsection{Depthwise Separable Convolution} \label{subsec:depthwiseseparableconvolution}

\begin{figure}
\centering
\includegraphics[width=8cm,keepaspectratio]{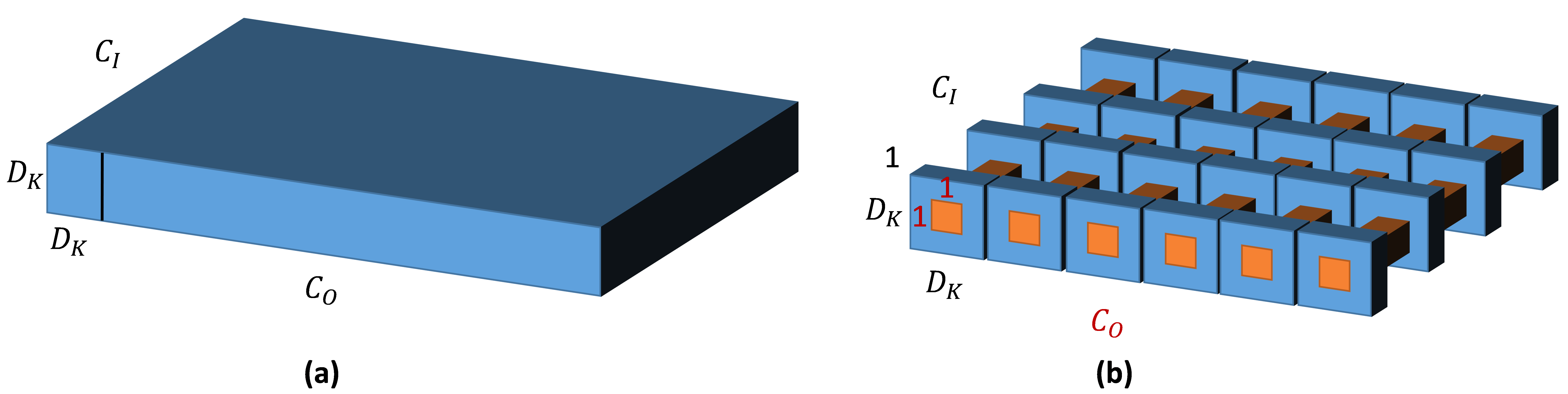}
\caption{(a)The general convolutional filter (b) The depthwise separable filter}
\label{fig:convolutionfilter}
\end{figure}

Before detailing our network architecture, we give a brief review of the depthwise seperable convolution in this section. First, we consider the computational complexity of a general convolution.
Let us denote the size of a general convolutional layer by $D_{K} \times D_{K} \times C_{I} \times C_{O}$, where $D_{K}$ is the size of kernel $K$, and $C_{I}$ and $C_{O}$ are the number of input and output channels, respectively. 
The dimension of input map is $W_{I} \times H_{I} \times C_{I}$, where $W_{I}$ and $H_{I}$ are the width and height of the input feature map, respectively. 
The size of output map $O$ is $W_{O} \times H_{O} \times C_{O} $, where $W_{O}$ and $H_{O}$ are the width and height of the map, respectively. 
Figure~\ref{fig:convolutionfilter}(a) shows a common feature convolutional layer. 
The computational cost of the common convolution layer is $W_{I} \times H_{I} \times C_{I} \times D_{K} \times D_{K} \times C_{O}$.



In depthwise separable convolution, we split the general convolution layer into two layers. 
One is the depthwise convolution layer with size $D_{K} \times D_{K} \times 1$ of a 2D kernel filter per each input channel $C_{I}$,  and the other is the pointwise convolution layer with  $1\times1$ convolution used to generate a linear combination of the output of the depthwise layer, as shown in Figure~\ref{fig:convolutionfilter}(b). 
The computational cost of the depthwise separable convolution layer can be derived by the following equation: $ W_{I} \times H_{I} \times C_{I} \times (  D_{K} \times D_{K} + C_{O} ) $.


Dividing the computational cost of general convolution by depthwise separable convolution, we can obtain the computational cost ratio is
$\frac{D_{K}^2 \times C_{O}}{D_{K}^2 + C_{O}}$. 
The greater the number of channels, the greater the speedup the depthwise separable convolution can be achieved.
In~\cite{howard2017mobilenets}, Howard~\etal demonstrate that simplifying the architecture of a CNN in this manner can considerably increase the inference speed without sacrificing the classification performance.


\subsection{Lightweight Multi-Task Network}

The Lightweight Multi-Task CNN (LMTCNN) is composed of one general convolution layer, two depthwise separable convolution layers and two fully connected layers. Thus, it can accomplish multiple tasks while reducing the memory cost. The system overview is shown in Figure~\ref{fig:lightenedmultitaskcnn}. To handle the age and gender classification on the Adience dataset~\cite{eidinger2014age}, our proposed method consists of three steps.

First, input color face image is scaled to the $256\times256$ and then cropped into the $227\times227$ in size using over-sampling. The over-sampling here means that the system extracts five cropped regions from the scaled color face image, four cropped regions from the corners and one from the center of the scaled color face image. The LMTCNN processes five cropped regions with their horizontal reflections and estimates the final result by the average score of these regions.

Second, the size $3\times7\times7$ pixel values of 96 kernel filters are applied to the input in the first general convolution layer, followed by a rectified linear unit (ReLU), a max pooling layer with window size equals to $3\times3$ and strides equal to two pixels and a local response normalization layer. The output feature map (size $28\times28\times96$)of the first general convolution layer is processed by the two subsequent depthwise separable convolution layers defined in Table~\ref{tab:depthwiseconvolutions1}. The output feature map of the last depthwise separable convolution layer is fed to the kernel size $3\times3$ of a max pooling layer that partitions the input feature map into a set of non-overlapping regions.

\begin{table}[]
\centering
\caption{The architecture of the two depthwise separable convolution layers used in the lightweight multi-task CNN}
\label{tab:depthwiseconvolutions1}
\begin{tabular}{@{}ccc@{}}
\toprule
Type      & Filter Shape        & Input Size     \\ \midrule
dw Conv1  & 3 x 3 x 96          & 28 x 28 x 96 \\
pw Conv1  & 1 x 1 x 96 x 256     & 28 x 28 x 96 \\
dw Conv2  & 3 x 3 x 256          & 14 x 14 x 256 \\
pw Conv2  & 1 x 1 x 256 x 384    & 14 x 14 x 256   \\ \bottomrule
\end{tabular}
\end{table}

Finally, the output feature map of the max pooling layer is fed to the two fully connected layers which contain 512 neurons, followed by a ReLU and a dropout layer. To achieve both the age classification for eight age classes and the gender classification for two gender classes, two separate softmax layers are followed by the output feature map of the average pooling layer. The first softmax layer assigns a probability for each class of the age and the other assigns a probability for each class of the gender. 
Figure~\ref{fig:lightenedmultitaskcnn} shows the network configuration visualization.

\section{Experiment and Result} \label{sec:experimentresult}
Our proposed network is implemented using the Tensorflow framework~\cite{abadi2016tensorflow}. Training and Testing are executed on the desktop with Intel Xeon E5 3.5 GHz CPU, 64G RAM and GeForce GTX TITAN X GPU. Training our proposed network takes approximately six hours. When testing on the desktop, predicting age and gender on a single image requires approximately $7.6$ milliseconds.

\subsection{The Result of Adience Dataset}
The Adience dataset~\cite{eidinger2014age} is composed of pictures taken by camera from smartphone or tablets. 
The images of Adience dataset capture extreme variations, including extreme blur (low-resolution), occlusions, out-of-plane pose variations, expressions. 
The entire Adience dataset includes 26,580 unconstrained images of 2,284 subjects. 
Its age labels contain eight groups, including $(0-2),(4-6),(8-13),(15-20),(25-32),(38-43),(48-53),(60+)$.

Unlike other datasets (such as Morph II) where the face images are taken in a controlled envoriment, the Adience dataset is a \emph{in-the-wild} benchmark for joint age and gender estimation, and is thus more demanding.
Because our purpose is to develop a mobile system that can estimate age and gender in real environments, testing this benchmark can reflect the performance more appropriately.


For age and gender classification, we measure and compare the accuracy using a five-fold cross validation. The number of images in each fold for training, validation and testing are shown in Table~\ref{tab:numberfivefolds}. The in-plane aligned version of the faces defined in~\cite{eidinger2014age} is used.

\begin{table}[]
\centering
\caption{The number of images in each fold of the training, validation and testing sets }
\label{tab:numberfivefolds}
\begin{tabular}{@{}lccc@{}}
\toprule
Fold   & Training  &  Validation & Testing \\ \midrule
First   & 11,136    &  1,242      & 3,879 \\
Second  & 11,905    &  1,348      & 3.005 \\
Third   & 11,814    &  1,323      & 3,121 \\
Forth   & 12,056    &  1,335      & 2,866 \\
Fifth   & 11,593    &  1,277      & 3,387 \\ \bottomrule
\end{tabular}
\end{table}

We compare our proposed method with baseline Levi\_Hassner CNN~\cite{levi2015age} by using five-fold cross validation with the number of images shown in Table~\ref{tab:numberfivefolds} to train by the training set and test by the testing set in each fold. Our proposed method is LMTCNN with the width multiplier of each depthwise separable convolution equals to 1 or 2. In Table~\ref{tab:agegenderaccuracy}, we demonstrate that the accuracy of the LMTCNN with $width~multiplier=2$ of the first depthwise separable convolution and $width~multiplier=1$ of the second depthwise separable convolution for age and gender classification.
As can be seen, although our architectures are more compact, their performance are comparable to that of the Levi\_Hassner CNN~\cite{levi2015age}.

\begin{table}[]
\centering
\caption{The accuracy of the age and gender classification generated by five-fold cross validation in Adience dataset }
\label{tab:agegenderaccuracy}
\begin{tabular}{@{}lccc@{}}
\toprule
& Age &  & Gender\\ \midrule
Methods  &  Top-1  &  Top-2 & Top-1  \\
 & Acc.(\%) & Acc.(\%) & Acc.(\%) \\ \midrule
Levi\_Hassner CNN~\cite{levi2015age} & 44.14 & 69.98 & 82.52 \\
LMTCNN-1-1 & 40.84 & 66.10 & 82.04 \\
LMTCNN-2-1 & \textbf{44.26} & \textbf{70.78} & \textbf{85.16} \\\bottomrule
\end{tabular}
\end{table}



\subsection{Mobile Applications}

To run a deep neural network model on mobile devices with Android operation system, we convert deep neural network model into the computational graph of Tensorflow library~\cite{abadi2016tensorflow} and we compare the model size of each method, as shown in Table~\ref{tab:modelsize}. The model size of LMTCNN with $width~multiplier=1$ of the first depthwise separable convolution and $width~multiplier=1$ of the second depthwise separable convolution is approximately nine times smaller than that of Levi\_Hassner CNN~\cite{levi2015age}, and the model size of LMTCNN with $width~multiplier=2$ of the first depthwise separable convolution and $width~multiplier=1$ of the second depthwise separable convolution is approximately half smaller.

\begin{table}[]
\centering
\caption{The Comparison of the model size}
\label{tab:modelsize}
\begin{tabular}{@{}lc@{}}
\toprule
Methods  & Model size (MB)   \\ \midrule
Levi\_Hassner CNN~\cite{levi2015age} & 70.8 \\
LMTCNN-1-1 &  \textbf{ 8.7 } \\
LMTCNN-2-1 & \textbf{ 30.0 } \\ \bottomrule
\end{tabular}
\end{table}

\begin{figure}
\centering
\includegraphics[scale=0.2]{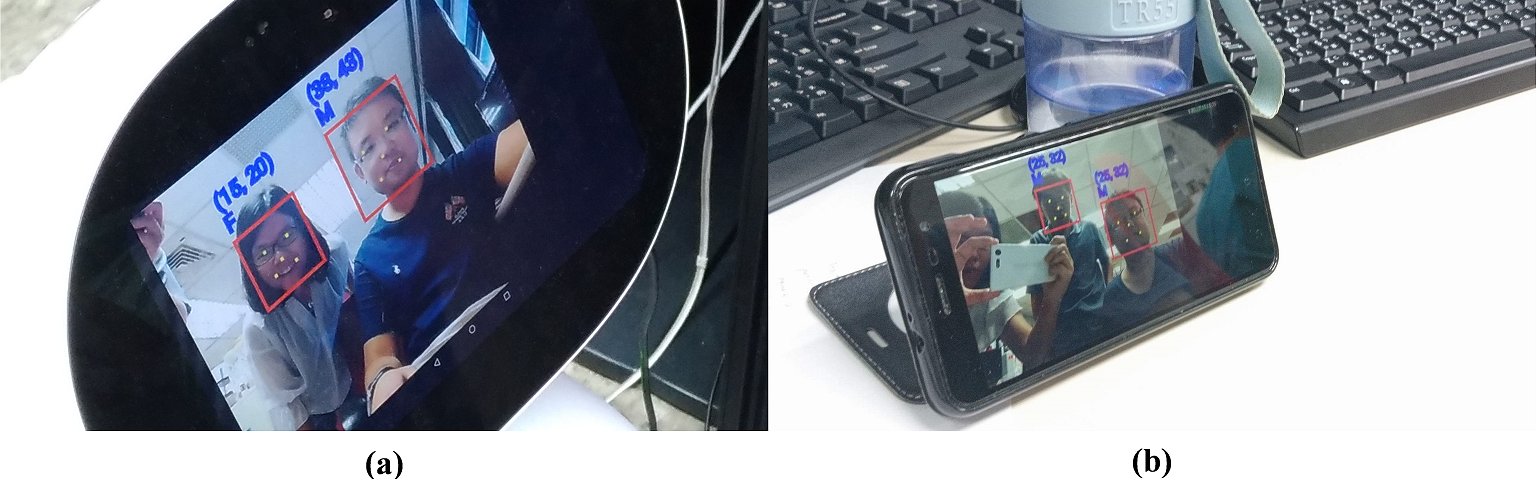}
\caption{The demonstration of LMTCNN executed on mobile devices. (a) Asus Zenbo and (b) Asus Zenfone 3.}
\label{fig:mobiledevice}
\end{figure}

For mobile application, we port the system with face detection, age recognition, and gender recognition on mobile devices, such as smartphone, tablet and smart robot. The face detection is implemented using the method of the MTCNN~\cite{zhang2016joint}. Then the region the facial regions are cropped from each frame for LMTCNN or Levi\_Hassner CNN~\cite{levi2015age} to recognize the age and gender. Figure~\ref{fig:mobiledevice} demonstrates our system on the ASUS Zenbo and ASUS Zenfone 3. The ASUS Zenbo is a smart robot developed by the ASUS incorporation with Intel Atom x5-Z8550 2.4GHz CPU, 4G RAM and Android 6.0.1 system and The ASUS Zenfone 3 is a smartphone developed by the ASUS incorporation with Qualcomm Snapdragon 625 2.02GHz CPU, 3G RAM and Android 7.0 system. We also calculate the processing time of each method on the ASUS Zenbo and ASUS Zenfone 3, as shown in Table~\ref{tab:mobilespeed}.

\begin{table}[]
\centering
\caption{The speed of each method executed in the mobile devices}
\label{tab:mobilespeed}
\begin{tabular}{@{}lcc@{}}
\toprule
Methods & Asus Zenbo & Asus Zenfone 3  \\
 & Speed & Speed \\
& (ms/frame) & (ms/frame) \\ \midrule
Levi\_Hassner CNN~\cite{levi2015age} & $\approx$ 4800 & $\approx$ 4800  \\
LMTCNN-1-1 & \textbf{ 204.7 } & \textbf{ 204.9 } \\
LMTCNN-2-1  & \textbf{ 297.6 } & \textbf{ 367.2 } \\\bottomrule
\end{tabular}
\end{table}


In summary, the above results (Table~\ref{tab:agegenderaccuracy},~\ref{tab:modelsize} and~\ref{tab:mobilespeed}) reveal that LMTCNN can decrease the size of model and speed up the inference on mobile devices while maintaining the accuracy of age and gender classification.

\section{Conclusion} \label{sec:conclusion}

We introduce the new network structure, LMTCNN, which accomplishes multiple tasks while maintaining the accuracy of age and gender classification.
We also show that our architecture can be realized on mobile devices with limited computational resources.
In the future, we will improve the performance of LMTCNN and reduce the size of model for the datasets of unconstrained face images with face attributes.

\bibliographystyle{latex8}
\bibliography{./bibs/agegender.bib,./bibs/deep.bib,./bibs/face.bib}

\end{document}